\documentclass[a4paper,conference]{IEEEtran}
\usepackage{graphicx}
\usepackage{cite}
\usepackage{amsmath,amssymb,amsfonts}
\usepackage[ruled,vlined]{algorithm2e}

\usepackage{graphicx}
\usepackage{textcomp}
\usepackage{xcolor}

\usepackage{fancyhdr}

\usepackage[utf8]{inputenc}		
\usepackage{color}				
\usepackage{microtype} 			
\usepackage{array}
\newcolumntype{C}[1]{>{\centering\arraybackslash}p{#1}}
\newcolumntype{L}[1]{>{\raggedright\arraybackslash}p{#1}}
\newcolumntype{R}[1]{>{\raggedleft\arraybackslash}p{#1}}
\usepackage{arydshln}

\ifCLASSINFOpdf
\else
\fi
\begin{document}
%
\title{An Investigation of Feature Selection and Transfer Learning for Writer-Independent Offline Handwritten Signature Verification}


\author{\IEEEauthorblockN{Victor L. F. Souza$^1$, Adriano L. I. Oliveira$^1$, Rafael M. O. Cruz$^2$ and Robert Sabourin$^2$}
\IEEEauthorblockA{$^1$\textit{Centro de Informática - Universidade Federal de Pernambuco}, Recife, Pernambuco, Brazil\\
Email: vlfs@cin.ufpe.br, alio@cin.ufpe.br}
\IEEEauthorblockA{$^3$\textit{École de technologie supérieure - Université du Québec}, Montreal, Quebec, Canada \\
Email: rafaelmenelau@gmail.com, robert.sabourin@etsmtl.ca}}


%


\maketitle

\begin{abstract}

\textit{SigNet} is a state of the art model for feature representation used for handwritten signature verification (HSV). This representation is based on a Deep Convolutional Neural Network (DCNN) and contains 2048 dimensions. When transposed to a dissimilarity space generated by the dichotomy transformation (DT), related to the writer-independent (WI) approach, these features may include redundant information. 
This paper investigates the presence of overfitting when using Binary Particle Swarm Optimization (BPSO) to perform the feature selection in a wrapper mode. 
We proposed a method based on a global validation strategy with an external archive to control overfitting during the search for the most discriminant representation.
Moreover, an investigation is also carried out to evaluate the use of the selected features in a transfer learning context. The analysis is carried out on a writer-independent approach on the CEDAR, MCYT and GPDS datasets.
The experimental results showed the presence of overfitting when no validation is used during the optimization process and the improvement when the global validation strategy with an external archive is used. Also, the space generated after feature selection can be used in a transfer learning context.

\end{abstract}



%
\IEEEpeerreviewmaketitle

\section{Introduction}

In the handwritten signature verification (HSV) problem, signature images are used to verify whether a person is whom he/she claims to be. In the offline context, the signature image is acquired after the writing process is completed \cite{hafemann_review:17}.

In general, two approaches are used for offline handwritten signature verification systems, writer-dependent (WD) and writer-independent (WI). In the WD scenario, a different classifier is trained for each writer and is responsible for verifying his/her signatures.
In the WI context, a single classifier is trained for all writers. In this case, the verification is carried out in the dissimilarity space resulting from the comparison between a questioned and reference signatures, through the Dichotomy Transformation (DT) \cite{hafemann_review:17}. 

The WI approach has the advantages of being scalable and adaptable, and can be used in a transfer learning context\cite{souza_ijcnn:19}, which is a methodology that tries to use the knowledge acquired from one task to solve related ones \cite{shao:15}. 

A state of the art Deep Convolutional Neural Network (DCNN) model for feature representation in the HSV context is the \textit{SigNet}, proposed by Hafemann et al. \cite{hafemann:17}. \textit{SigNet} feature vectors are composed of 2048 dimensions. However, when using this representation in a WI context, some of the features may be redundant and have little importance in the generated dissimilarity space. Thereby, swarm optimization algorithms can be used for feature selection to obtain only the relevant dimensions on the transposed space \cite{cruz:17}.

We propose to use a feature selection technique based on binary particle swarm optimization (BPSO) for WI handwritten signature verification. The optimization is conducted based on the minimization of the error of the WI classifier in a wrapper mode. The wrapper approach evaluates a specific machine learning algorithm to find optimal features and is susceptible to overfitting \cite{radtke:06}. 
In wrapper based feature selection techniques, the optimization process becomes another learning process that is subject to overfit the training data. This happens when the optimized feature set memorizes the training data instead of producing a general model \cite{dossantos:09}.
In the literature is known that a validation strategy can be used during the optimization process to control the overfitting \cite{cruz:17}.
In this study we analyse two strategies, (i) partial validation and (ii) global validation with external archive strategy, to check whether they can control overfitting and consequently improve BPSO-based feature selection performance.


Thus, the objectives of this study are: (i) to investigate the presence of redundant information in the dissimilarity space generated by DT. (ii) Whether the overfitting control during the optimization helps in having a feature representation that can generalize better across different datasets. (iii) Whether this space generated after feature selection can be used in a transfer learning context.

This paper is organized as follows: Section \ref{sec:basic_concepts} contains the basic concepts related to this work. In Section \ref{sec:feature_selection} how the binary particle swarm optimization (BPSO) can be used for feature selection together with the used fitness function and the proposed overfitting control approach. Section \ref{sec:experiments} contains the experiments and the discussion about the obtained results. In the last section, the conclusion and future works are presented.

\section{Basic concepts}
\label{sec:basic_concepts}

\subsection{Handwritten Signature Verification (HSV)}

The key task for an HSV system is deciding whether a given signature image is genuine or a forgery. Intuitively genuine signatures are those that belong to the indicated person and forgeries are those created by someone else \cite{hafemann_review:17}. 

Forgeries, in general, are divided into \cite{hafemann_review:17}: (i) Random forgeries: the forger neither knows the name nor the signature pattern of the original writer. (ii) Skilled forgeries: the forger knows both the name and the signature pattern of the original writer, which results in better forgeries.

Systems that deal with the offline HSV can be divided into Writer-Dependent (WD) or Writer-Independet (WI) systems. While in the first case, a classifier is trained for each writer, in WI systems a single model is trained for all writers from a dissimilarity space generated by the dichotomy transformation (DT) \cite{cha:00}. In DT a dissimilarity (distance) measure is used to compare two samples as belonging to the same writer or not \cite{eskander:13}. 
Hence, the model can verify signatures of writers for whom the classifier was not trained.
When compared to the WD approach, WI systems have the advantages of being less complex and more scalable, but in general, obtain worse accuracy \cite{hafemann_review:17}.

\subsection{Feature representation}
\label{sec:feature_representation}

The \textit{SigNet}, proposed by Hafemann et al. \cite{hafemann:17}, is a state of the art Deep Convolutional Neural Network (DCNN) model for feature representation in the HSV problem. 
The idea of this approach is to cluster different writers in separate regions of the new representation space, based on the most representative properties of their signatures. Table \ref{tab:dcnn_summary} summarizes \textit{SigNet} architecture.

\begin{table}[!htb]
\caption{Summary of the \textit{SigNet} layers}
\label{tab:dcnn_summary}
\scriptsize
\centering

\begin{tabular}{lll}
\hline
Layer & Size & Other Parameters \\ 
\hline
Input & 1 x 150 x 220 & \\
Convolution (C1) & 96 x 11 x 11 & Stride = 4, pad = 0 \\
Pooling & 96 x 3 x 3 & Stride = 2 \\
Convolution (C2) & 256 x 5 x 5 & Stride = 1, pad = 2 \\
Pooling & 256 x 3 x 3 & Stride = 2 \\
Convolution (C3) & 384 x 3 x 3 & Stride = 1, pad = 1 \\
Convolution (C4) & 384 x 3 x 3 & Stride = 1, pad = 1 \\
Convolution (C5) & 256 x 3 x 3 & Stride = 1, pad = 1 \\
Pooling & 256 x 3 x 3 & Stride = 2 \\
Fully Connected (FC6) & 2048 & \\
\textbf{Fully Connected (FC7)} & \textbf{2048} & \\
Fully Connected + Softmax ($P(\textbf{y}|X)$) & M & \\
\hline

\end{tabular}
\end{table}

An important aspect of \textit{SigNet} is that it works in a writer-independent way.
For new writers, \textit{SigNet} is used to project the signature images onto the new representation
space, by using feed-forward propagation until the FC7 layer, obtaining feature
vectors with 2048 dimensions. Thus, it is scalable for new incoming writers, different from those in which it was trained.

In this work, our original feature space is represented by these 2048 features \cite{hafemann:17} (which are available online\footnote{http://en.etsmtl.ca/Unites-de-recherche/LIVIA/Recherche-et-innovation/Projets/Signature-Verification}). 
However, many of these features may be redundant in the transposed dissimilarity space (generated by DT) and may have little influence on distinguishing between positive and negative samples. 
So, feature selection techniques can be used for both obtain only those features that are relevant and further improve the performance of the model used for verification purposes.

\subsection{WI dichotomy transformation}

The Dichotomy Transformation (DT), proposed by Cha and Srihari \cite{cha:00}, transforms a multi-class pattern recognition problem into a $2$-class problem. 
In this approach, a dissimilarity (distance) measure is used to distinguish whether a given reference and a questioned sample belong to the same class or not \cite{costa:19}. When applied to the handwritten signature verification (HSV) context, it characterizes the writer-independent (WI) approach, the samples are signatures and to perform the verification means belonging to the same writer or not \cite{bertolini:16}. 

Formally, given the images of a questioned signature $I_q$ and a reference signature $I_r$, the first step is to extract the feature vectors $\textbf{x}_q$ and $\textbf{x}_r$, respectively. In our case, we use \textit{SigNet} for that. Then, the dissimilarity vector resulting from the Dichotomy Transformation, $\textbf{u}$, is computed by equation \ref{eq:DT_distance}:

\begin{equation}
\label{eq:DT_distance}
  \textbf{u}(\textbf{x}_{q},\textbf{x}_{r}) =
  \begin{bmatrix}
    |x_{q1} - x_{r1}| \\
    |x_{q2} - x_{r2}| \\
    \vdots \\
    |x_{qn} - x_{rn}|
  \end{bmatrix}
\end{equation}

\noindent where $| \cdot |$ represents the absolute value of the difference, $x_{qi}$ and $x_{ri}$ are the $i$-th features of the signatures $\textbf{x}_q$ and $\textbf{x}_r$ respectively, and $n$ is the number of features. Hence, each dimension of the $\textbf{u}$ vector is equal to the distance between the corresponding dimensions of the vectors $\textbf{x}_q$ and $\textbf{x}_r$, and therefore all these vectors have the same dimensionality \cite{bertolini:16}.

As mentioned, regardless of the number of writers, after applying DT, only two classes are present in the dissimilarity space:
\begin{itemize}
    \item The \textit{within/positive class ($u_+$)}, when the reference and questioned feature vectors used to obtain the dissimilarity vector belong to the same writer.
    \item The \textit{between/negative class ($u_-$)}, otherwise.
\end{itemize}

Once the data is transposed into the dissimilarity space, a dichotomizer (i.e., a 2-class classifier) is trained and used to perform the verification task.
It is expected the trained dichotomizer to be able to distinguish if two samples belong to the same writer or not \cite{cha:00}.

\subsection{Transfer Learning}

Transfer learning (TL) techniques aim to extract useful information from a source domain and apply it to a target domain \cite{shao:15}.

As mentioned, the WI verification only depends on the reference signature used as input to the classifier. So, by using the DT in a writer-independent approach, the dichotomizer can verify signatures of writers for whom the classifier was not trained (i.e., transfer learning). Consequently, a single model already trained can be used to verify the signatures of new incoming writers, without requiring additional training or updating of the model.

In the WI-HSV context, Souza et al. \cite{souza_ijcnn:19} showed that a WI-SVM trained in the GPDS dataset was able to verify signatures obtained from other datasets without any further transfer adaptation. 
It is worth noting that different databases have different acquisition protocol (scanner,
writing space, writing tool etc).
Still, the WI-SVM obtained similar results when compared to both WD and WI classifiers trained and tested in their own datasets.

\section{Feature selection using binary particle swarm optimization (BPSO)}
\label{sec:feature_selection}

The objective of feature selection techniques is to identify the most relevant subset of features from the entire set of features considered. 
The motivations for using this approach include reduction of the computational complexity, reduction of dimensionality, removal of non-informative features, enhanced generalization power by reducing overfitting \cite{cruz:17}.

In the context of feature selection, particle swarm optimization algorithms are used in their binary version (BPSO) and have been obtaining good results when compared to other optimization algorithms used for this task \cite{chuang:11}. 

For a formal definition, given a binary search space with $D$ dimensions and a swarm with $N$ particles, the $i$-th particle of the swarm can be represented by a $D$-dimensional vector $\textbf{x}_i = [x_{i1}, x_{i2}, ... , x_{iD}]$, which corresponds to the position of the particle in space. In this work context, each dimension $x_{id}$ represents a single feature and value ``1'' means that the respective feature is selected and ``0'' otherwise.
The particle velocity consists of $\textbf{v}_i = [v_{i1}, v_{i2}, ..., v_{iD}]$; the best position found by the particle as $\textbf{pBest}_i = [pBest_{i1}, pBest_{i2}, ..., pBest_{iD}]$ and the best position obtained by the swarm as $\textbf{gBest} = [gBest_{1}, gBest_{2}, ..., gBest_{D}]$. Then, for each iteration, the update of the velocity and the position occurs respectively by the equations \ref{eq:velocity} and \ref{eq:position} \cite{zhang:13}.

\begin{equation}
\begin{split}
\label{eq:velocity}
  \textbf{v}_{i}(t + 1) & = w \cdot \textbf{v}_{i}(t) + c_1 \cdot rand \cdot (\textbf{pBest}_i - \textbf{x}_{i}(t)) \\ 
  & + c_2 \cdot Rand \cdot (\textbf{gBest}_{i} - \textbf{x}_{i}(t))
\end{split}
\end{equation}

\begin{equation}
\label{eq:position}
  \textbf{x}_{i}(t + 1) =
  \begin{cases}
    \textbf{x}_{i}(t)^{-1}\:\: If \:\: rand_{p} < T(\textbf{v}_{i}(t + 1)) \\
    \textbf{x}_{i}(t)     \:\:\:\:\:\: If \:\: rand_{p} \geq T(\textbf{v}_{i}(t + 1)) \\
  \end{cases}
\end{equation}

\noindent where, {\it $c_1$} and {\it $c_2$} represent acceleration factors and are positive constants; $rand$, $Rand$ and $rand_{p}$ are random variables with uniform distribution within the interval $[0,1]$, and {\it w} is the weight of inertia. In the velocity equation, the first factor represents inertia, the second factor the cognitive component and the third factor the social component.

As we are dealing with a binary search space, updating the position of a particle means switching between selecting the feature (``1'') or not (``0''). 
The transformation of the continuous search space into a binary space is conducted by using a transfer function, $T$ \cite{mirjalili:13}.

Mirjalili et al. \cite{mirjalili:13} bring that the choice of a well-suited transfer function is crucial in order to obtain good convergence performance. According to the authors, in general, the V-Shaped transfer functions present better behavior both in terms of avoiding local minima and convergence speed \cite{mirjalili:13}.

\begin{equation}
\label{eq:v_shape}
  T(x) =
  \begin{vmatrix}
    \frac{2}{\pi} arctan (\frac{\pi}{2}x)
  \end{vmatrix}
\end{equation}

An important aspect is that the V-Shaped transfer functions encourage particles to stay in their current positions when their velocity values are low or switch to their complements when the velocity values are high. The V-Shaped transfer function is used in this work and can be computed through equation \ref{eq:v_shape}.

Research has shown that the three parameters {\it w}, {\it $c_1$} and {\it $c_2$} have a significant impact on the algorithm performance \cite{hassan:05}. Thus, we choose to use Improved Self-Adaptive Particle Swarm Optimization Algorithm (IDPSO) \cite{zhang:13}. In this variation of PSO, the algorithm itself adjusts {\it w}, {\it $c_1$} and {\it $c_2$} dynamically over iterations, promoting global search in the beginning and local search in the final iterations.

\subsection{Fitness function}

We propose to use a BPSO-based feature selection for WI handwritten signature verification with the fitness function as being the WI classifier performance in wrapper mode. 
In wrapper methods, a predictive model is used to evaluate a combination of features and assign a score based on model accuracy \cite{radtke:06}.

The optimization is conducted based on the minimization of the Equal Error Rate ($EER$) of the SVM in a wrapper mode. 
The $EER$ metric is the error obtained when False Rejection Rate ($FRR$) is equal to False Acceptance Rate ($FAR$) \cite{souza:18}. The user threshold (considering just the genuine signatures and the skilled forgeries) was employed \cite{souza:18}.
The motivation for using Support Vector Machines (SVM) as the classifier is because it is one of the most effective classifiers for both writer-dependent (WD) and writer-independent (WI) signature verification tasks \cite{hafemann_review:17}.

\subsection{Overfitting control approach}

In the feature selection scenario, overfitting occurs when the optimized feature set memorizes the training set instead of producing a general model. Hence, it may fail to generalize well to unseen data.
To decrease the chance of overfitting, a validation procedure can be used during the optimization process in order to select solutions with good generalization.

According to Santos et al. \cite{dossantos:09}, one possible approach is the partial validation (\textit{PV}) strategy where the validation occurs only at the end of the optimization process by validating the last population on another set of unknown observations – the selection set. By using this approach, the optimization routine is more likely to produce better results than selecting solutions based solely on the accuracy of the optimization set alone. However, this strategy has the disadvantage that the solution is validated only once, after the optimization process is completed.

We propose to use a global validation (\textit{GV}) strategy, where the validation of the candidate solutions is executed in all iterations of the optimization process. This can be accomplished by storing the best solutions in an external archive \cite{radtke:06}.

\begin{algorithm}[!htb]
\small
\SetAlgoLined
\KwResult{External archive $A$}
 Create initial population $P(1)$ with $N$ individuals\\
 Replace optimization set by the selection set for objective function evaluation\\
 Calculate objective functions for all solutions in $P(t)$\\
 $A = \emptyset$\\
 $t = 1$\\
 \While{t $<$ maximum iterations}{
    Evolve $P(t)$ to $P(t+1)$\\
    Validate $P(t+1)$ with the selection set\\
    Update the external archive $A$ with the individuals from $P(t+1)$ based on their fitness from the validation process\\
    t=t+1\\
 }
 \caption{Pseudo-code showing the global validation strategy with external archive to control overfitting.}
 \label{alg:external_archive}
\end{algorithm}

Algorithm \ref{alg:external_archive} presents the pseudo-code for the global validation strategy. As can be seen, an empty external archive $A$ is created at the beginning and updated at each iteration according to the validated solutions.
During this routine, the optimization set ($Opt$) is temporarily replaced by the selection set ($Sel$) to evaluate the fitness function. 
At each iteration, all the best solutions previously found are grouped with the population of the new swarm and then ranked. Finally, the external archive maintains the $N$ best candidate solutions.

\begin{figure}[!htb]
\centering
  \includegraphics[width=\columnwidth]{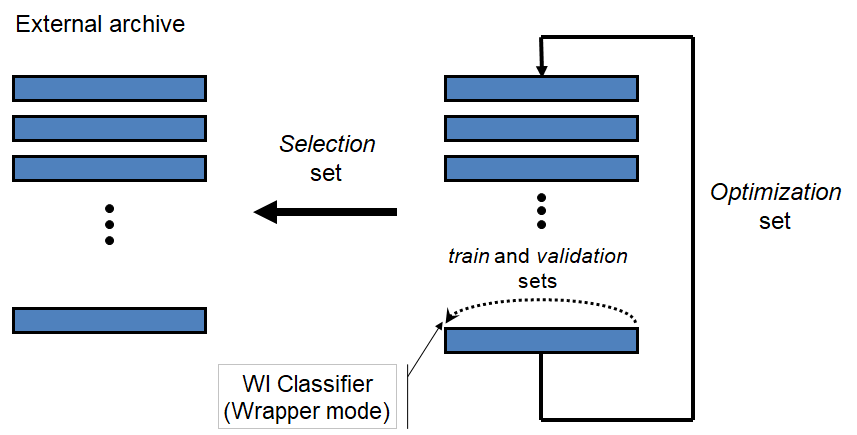}
  \caption{Global validation with eternal archive strategy overview.}
  \label{fig:gecco_overview}
\end{figure}

Figure \ref{fig:gecco_overview} depicts the global validation strategy overview. The WI classifier is used in a wrapper mode considering the train and validation sets. The optimization set ($Opt$) is used to guide the search during the iterations of the BPSO. In turn, the Selection set ($Sel$) is used in the validation stage for any of the methods used to control the overfitting.

\section{Experiments}
\label{sec:experiments}

The objectives of the experiments are to investigate both the use of feature selection through BPSO and the effectiveness of the overfitting control strategy. Thus, the experiments are conducted for four different situations: 
(i) the model without feature selection (i.e., the 2048 features are used) \cite{souza_ijcnn:19};
(ii) the model with feature selection and no validation scheme during the optimization (referred to as $BPSO_{NV}$ in this paper);
(iii) the partial validation model (referred to as $BPSO_{PV}$);
(iv) the global validation model (referred to as $BPSO_{GV}$) .

The same analysis is also carried out to check whether the space generated by the feature selection can be used in a transfer learning context.

\subsection{Dataset}

\begin{figure}[!htb]
\centering
  \includegraphics[width=7cm]{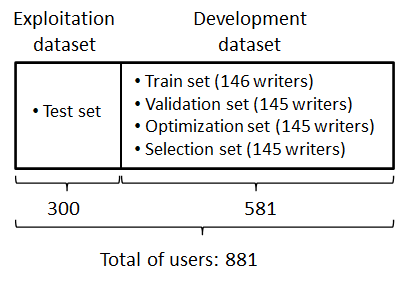}
  \caption{GPDS-300 dataset segmentation}
  \label{fig:gecco_dataset}
\end{figure}

In the experiments, the whole set of steps (training, feature selection and testing) are carried out using GPDS dataset, specifically in the GPDS-300 stratification \cite{hafemann_review:17}. MCYT and CEDAR datasets are considered only for test purposes on the transfer learning scenario.
Figure \ref{fig:gecco_dataset} depicts the segmentation of the writers on the GPDS-300 dataset.

As can be seen in Figure \ref{fig:gecco_dataset}, (i) the Exploitation set, where the tested set is acquired, is composed of writers 1 to 300. 
(ii) The Development set is formed by the other 581 writers: 146 writers are randomly selected to compose the train set ($train$), another 145 for the validation set ($Val$), another 145 for the optimization set ($Opt$) and the remaining 145 for the selection set ($Sel$).

\begin{table}[!htb]
\caption{Exploitation set $\varepsilon$}
\label{tab:summary_exploitation_set}
\footnotesize
\centering

\begin{tabular}{ccc}
\hline
Dataset & \#Writer & \#questioned signatures (per writer)\\ 
\hline
GPDS-300 & 300 & 10 genuine, 10 skilled  \\
MCYT & 75 & 5 genuine, 15 skilled  \\
CEDAR & 55 & 10 genuine, 10 skilled \\

\hline
\end{tabular}
\end{table}

Table \ref{tab:summary_exploitation_set} summarizes the used Exploitation set $\varepsilon$ for each dataset. It contains the number of writers used and the number of signatures for each writer, considering each of the test datasets. As the MCYT and the CEDAR datasets are considered only on the transfer learning scenario, all the writers belong to the exploitation set.

\subsection{Experimental setup}

The Equal Error Rate ($EER$) metric, using user thresholds (considering just the genuine signatures and the skilled forgeries) was used in the evaluation of the verification models \cite{hafemann:17}. 

In this paper, the SVM is used as writer-independent classifier with the following settings: $RBF$ kernel, $\gamma = 2^{-11}$ and $C=1.0$. The signed distance of the samples to the classifier's hyperplane is used as classifiers output \cite{souza:20}. 
Signatures were randomly selected, and a different SVM was trained for each replication (five replications were performed for each experimental configuration).

A total of 12 references per writer were considered.
The DT generates a different dissimilarity vector from each of these references which are passed down to the WI classifier and then the MAX of the signed distance is used as the partial decision fusion function \cite{souza_iccs:19}. 
In the training step (training and validation sets), the model uses genuine signatures and random forgeries. For each writer, 10 genuine signatures and 10 random forgeries are used as questioned signatures to obtain respectively the positive samples and the negative samples.
In its turn, during optimization (optimization and selection sets), the proposed approach needs genuine signatures and skilled forgeries. As mentioned, the fitness function minimizes the $EER$ with the user threshold considering only genuine signatures and skilled forgeries. In this case, for each writer, 10 genuine signatures and 10 skilled forgeries are used.

The paper by Souza et al. \cite{souza_ijcnn:19} showed that many of the samples generated by the dichotomy transformation are redundant and so using a prototype selection technique, such as the Condensed Nearest Neighbors (CNN) \cite{hart:68}, it is possible to speed up the classifier training and still achieve a classification performance that is similar to or better than what is obtained by using all the training samples.
Thus, all the operations carried out in this paper are performed in the space with reduced samples, i.e., after prototype selection through CNN. 

The IDPSO parameters are set to their default values, as presented by Zhang et al. \cite{zhang:13}. The population size is equal to 20, the acceleration constants $c_1 = c_2= 2.0$, $w_{inicial} = 0.9$ , $w_{final} = 0.4$ and $\mu = 100$. The maximum number of iterations was set to 40. 

\subsection{Overfitting analysis}

\begin{figure*}[!htb]
  \includegraphics[width=\textwidth]{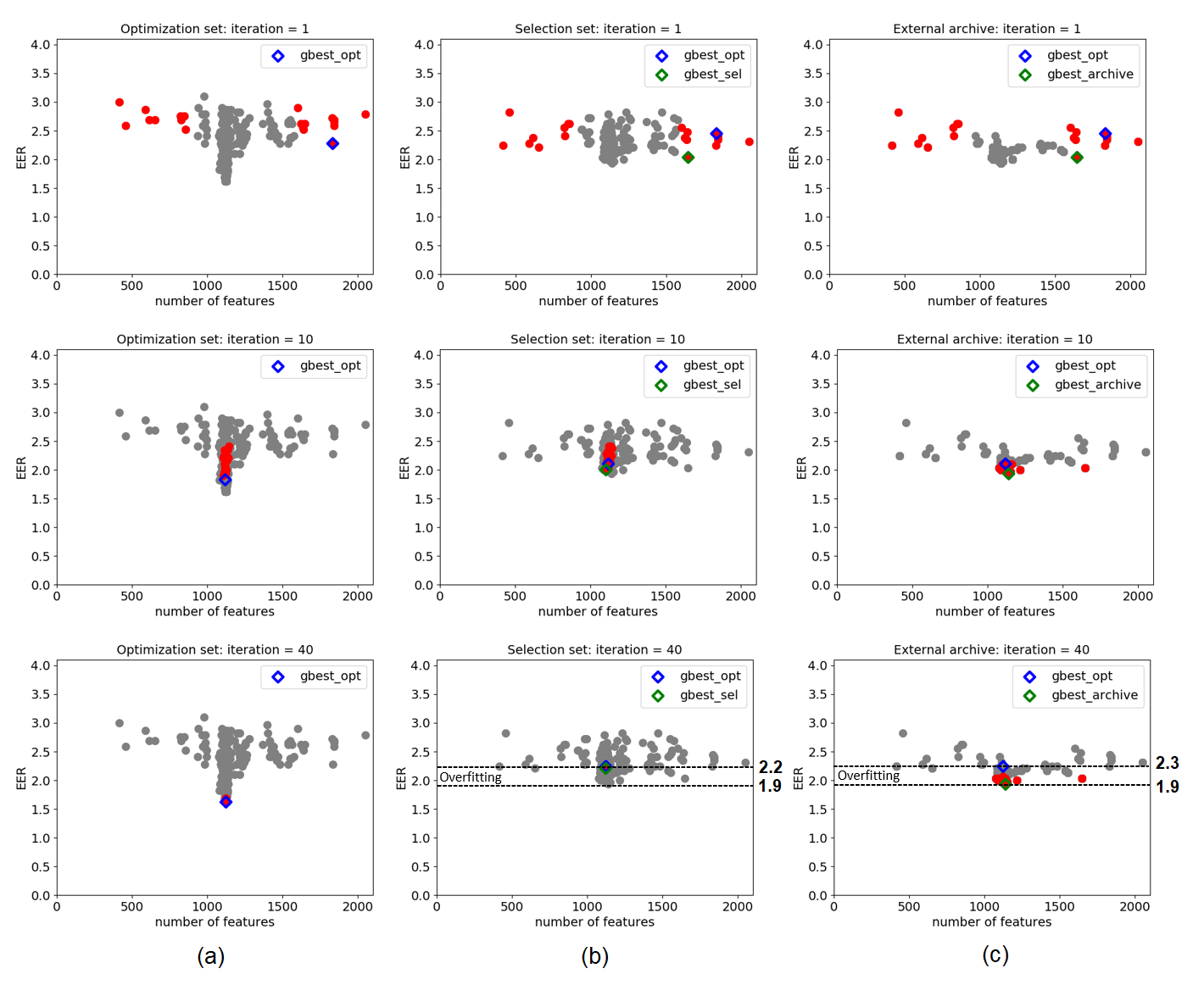}
  \caption{At first column (a) swarm behavior on the optimization set; in the second column (b) the swarm behavior when projected on the selection set; and in the third column (c) the swarm behavior in the external archive.}
  \label{fig:space_comparison_overfitting}
\end{figure*}

Figure \ref{fig:space_comparison_overfitting} depicts the convergence of the swarm (iterations 1, 10 and 40 are presented). 
Gray circles represent the whole set of candidate solutions, considering all iterations. Red circles represent the particles in the respective iteration. Blue diamond represents the information of the gbest from the optimization set. The green diamond represents the information of the best solution found in the selection set. 

We initialize half of the particles randomly in the intervals between [500, 1000] selected features and the other half in the intervals between [1500, 2048]. The objective was to extend the search space as much as possible. However, the particles soon converge into space containing around half the maximum number of features (i.e., 1024).

Column (a) details algorithm convergence during the optimization process, considering the own optimization set. 
In column (b), the same solutions are projected on the selection set ($BPSO_{PV}$ approach). 
Finally, column (c) simulates the convergence in the external archive obtained by projecting all the candidate solutions on the selection set at each generation $t$ ($BPSO_{GV}$ approach).

As can be seen in the second column of Figure \ref{fig:space_comparison_overfitting}, considering iteration 40, the overfitting truly happened when solutions are validated only at the last iteration ($BPSO_{PV}$). The estimated overfitting is about 0.3 $EER$ when compared to the lowest error rate in the external archive.
The second column also indicates that some candidate solutions that perform well in the selection set are discarded by the algorithm. This observation also confirms the needing for a validation stage at each iteration $t$. 

As depicted in the third column of Figure \ref{fig:space_comparison_overfitting}, considering iteration 40, we can also see the overfitting happening when solutions do not use any validation stage. The amount of overfitting measured from the best candidate solution, from the optimization set, is about 0.4 $EER$ when compared to the lowest $EER$ in the external archive. 

\subsection{Results and discussions}
\label{sec:gpds}

Table \ref{tab:models_comparison} presents the results obtained by the models. As can be seen, the overfitting which we already observed in Figure \ref{fig:space_comparison_overfitting} resulted in a worse $EER$ when compared to the scenario without feature selection, 3.76 against 3.47.

\begin{table}[!htb]
\caption{Comparison of $EER$ considering the presented models, in the GPDS-300 dataset (errors and standard deviations in \%)}
\label{tab:models_comparison}
\footnotesize
\centering

\begin{tabular}{ccc}
\hline
Approach & \#features & $EER$ \\ 
\hline
No feature selection & 2048 & 3.47 (0.15) \\ 
$BPSO_{NV}$ & 1124 & 3.76 (0.07) \\ 
$BPSO_{PV}$ & 1120 & 3.64 (0.08) \\ 
$BPSO_{GV}$ & 1140 & \textbf{3.46 (0.08)} \\ 
 \hline
\end{tabular}
\end{table}

In terms of overfitting control,  results indicate that the $BPSO_{NV}$ scheme is worse than the $BPSO_{PV}$, which in turn is worse than using the $BPSO_{GV}$ approach.
Thus, by using the global validation strategy, it is possible to control the overfitting of the model and, thereby, improve the performance of the BPSO-based feature selection approach.

Moreover, we can also see the presence of redundant features in the dissimilarity space generated by the dichotomy transformation. Notice that the $BPSO_{GV}$ model uses only 55\% of the total number of features and still manages to obtain a similar $EER$ when compared to the model trained with all the 2048 features.

Table \ref{tab:state_gpds_experiments4} contains the comparison of the presented models with the state of the art methods for the GPDS-300 dataset. Souza et al. \cite{souza:20} represents the WI-SVM trained in the original feature space.


\begin{table}[!htb]
\caption{Comparison of $EER$ with the state of the art, in the GPDS-300 dataset (errors and standard deviations in \%)}
\label{tab:state_gpds_experiments4}
\scriptsize
\centering

\begin{tabular}{ccccc}
\hline
Type & HSV Approach & \#Ref & \#Models & $EER$ \\ 
\hline
WD & Hafemann et al. \cite{hafemann:16} & 12  &  300 & 12.83 \\ 
WD & Zois et al. \cite{zois:16} & 5 & 300 &  5.48 \\ 
WD & Hafemann et al. (fine-tuned) \cite{hafemann:17} & 12 & 300 & 3.15 (0.18) \\ 
WD & Hafemann et al. (fine-tuned) \cite{hafemann:18} & 12  & 300 & 0.41 (0.05) \\ 
WD & Yilmaz and Ozturk \cite{yilmaz:18} & 12  & 300 & 0.88 (0.36) \\ 
WD & Zois et al. \cite{zois:19} & 12 & 300 & 0.70 \\ 
WI & Kumar et al. \cite{kumar:12} & 1 & 1 &  13.76 \\ 
WI & Eskander et al. \cite{eskander:13} & 1 & 1 &  17.82 \\ 
WI & Hamadene and Chibani \cite{hamadene:16} & 5 & 1 &  18.42 \\ 
WI & Zois et al. \cite{zois:19asymmetric} & 5 & 1 & 3.06 \\ 
WI & Souza et al. \cite{souza:20} & 12 & 1 & 3.47 (0.15) \\ 
\hdashline
WI & $BPSO_{GV}$ & 12 & 1 & 3.46 (0.08) \\ 

\hline
\end{tabular}
\end{table}

In general, our $BPSO_{GV}$ approach obtains a lower $EER$. In the WI scenario, it was able to outperform almost all the other methods. However, it presents similar results to Souza et al. \cite{souza:20} and is worse when compared to the model proposed by Zois et al. \cite{zois:19asymmetric}. 

Comparing with WD models, our approach was outperformed by Hafemann et al. \cite{hafemann:18} (fine-tuned), Yilmaz, and Ozturk \cite{yilmaz:18} and Zois et al. \cite{zois:19}, being better or comparable than the other methods.
It is important to point out that, as a WI model, our approach has greater scalability than these other models, since only one classifier is needed to perform signature verification.

\subsection{Transfer learning analysis}

In the paper by Souza et al. \cite{souza_ijcnn:19}, the authors experimentally showed that a WI-SVM trained in the GPDS can be employed to verify signatures in the other datasets without any further transfer adaptation. Herein, we investigate whether the space generated by the feature selection approach can also be used in a transfer learning context.

Table \ref{tab:models_comparison_transfer_learning} shows the results obtained when the models from Table \ref{tab:models_comparison}, trained in the GPDS dataset, are used to perform the verification in the CEDAR and MCYT databases. 

\begin{table}[!htb]
\caption{Comparison of $EER$ considering the presented models, in a transfer learning conxtext in the CEDAR and MCYT datasets (errors and standard deviations in \%)}
\label{tab:models_comparison_transfer_learning}
\scriptsize
\centering

\begin{tabular}{cccc}
\hline
Approach & \#features & $EER_{CEDAR}$ & $EER_{MCYT}$ \\ 
\hline
No feature selection & 2048 & 3.32 (0.22) & 2.89 (0.13) \\ 
$BPSO_{NV}$ & 1124 & 4.00 (0.17) & 2.69 (0.13) \\ 
$BPSO_{PV}$ & 1120 & 3.98 (0.25) & 2.56 (0.05) \\ 
$BPSO_{GV}$ & 1140 & \textbf{3.27 (0.22)} & \textbf{2.48 (0.23)} \\ 
 \hline
\end{tabular}
\end{table}

As can be seen, in terms of overfitting control for the transfer learning scenario, results indicate that the $BPSO_{NV}$ scheme is worse than the $BPSO_{PV}$, which in turn is worse than using the $BPSO_{GV}$ approach. Thus, by using the global validation strategy, it is possible to control the overfitting of the model and, thereby, improve the performance of the BPSO-based feature selection approach.

In the transfer learning scenario, the overfitting control has a higher impact when compared to the GPDS dataset, in which all training was conducted. The $BPSO_{NV}$ model tends to lose generalization performance in other datasets as the feature representation probably become too specialized to the overfitted data, and does not work well to out-of-distribution test. Thus, the overfitting control scheme helps in having a feature representation that can generalize better across different datasets as it does not specialize too much on the optimization one.

For both CEDAR and MCYT datasets, the $BPSO_{GV}$ approach presented the best results.
In the CEDAR dataset, the lack of generalization power for the $BPSO_{NV}$ and $BPSO_{PV}$ approaches resulted in worse results when compared to the model without feature selection. 
In its turn, for the MCYT dataset, all models with feature selection obtained better results when compared to the one using the whole set of features, the model with global validation being the best. Recall that these models with feature selection use only 55\% of the total number of features.

\begin{table}[!htb]
\caption{Comparison of $EER$ with the state of the art in the CEDAR dataset (errors and standard deviations in \%)}
\label{tab:state_cedar}
\scriptsize
\centering
\begin{tabular}{ccccc}
\hline
Type & HSV Approach & \#Ref & \#Models & $EER$ \\ 
\hline
WD & Okawa \cite{okawa:16} & 16 & 55 & 1.60  \\ 
WD & Serdouk et al. \cite{serdouk:16} & 16 & 55 & 3.52  \\
WD & Zois et al. \cite{zois:16} & 5 & 55 & 4.12  \\ 
WD & Hafemann et al. \cite{hafemann:17} & 12 & 55 & 4.76 (0.36) \\ 
WD & Zois et al. \cite{zois:17} & 5 & 55 & 2.07  \\ 
WD & Hafemann et al. (fine-tuned) \cite{hafemann:18} & 10 & 55 &  2.33 (0.88)\\ 
WD & Okawa \cite{okawa:18} & 16 & 55 & 1.00  \\ 
WD & Tsourounis et al. \cite{tsourounis:18} & 5 & 55 & 2.82  \\
WD & Zois et al. \cite{zois:18} & 5 & 55 & 2.30  \\
WD & Zois et al. \cite{zois:19} & 10 & 55 & 0.79  \\
WI & Chen and Srihari \cite{chen:06} & 16 & 1 & 7.90  \\ 
WI & Kumar et al. \cite{kumar:10} & 1 & 1 & 11.81 \\ 
WI & Kumar et al. \cite{kumar:12} & 1 & 1 &  8.33 \\ 
WI & Kumar and Puhan \cite{kumar:14} & 16 & 1 & 6.02  \\ 
WI & Guerbai et al. \cite{guerbai:15} & 12 & 1 & 5.60  \\
WI & Hamadene and Chibani \cite{hamadene:16} & 5 & 1 & 2.11  \\
WI & Zois et al. \cite{zois:19asymmetric} & 5 & 1 & 2.90  \\
WI & Souza et al. \cite{souza:20} & 12 & 1 & 3.32 (0.22) \\
\hdashline
WI & $BPSO_{GV}$ & 12 & 1 & 3.27 (0.22) \\ 
\hline
\end{tabular}
\end{table}

\begin{table}[!htb]
\caption{Comparison of $EER$ with the state of the art in the MCYT dataset (errors and standard deviations in \%)}
\label{tab:state_mcyt}
\scriptsize
\centering
\begin{tabular}{ccccc}
\hline
Type & HSV Approach & \#Ref & \#Models & $EER$ \\ 
\hline
WD & Vargas et al. \cite{vargas:11} & 10  & 75 &  7.08 \\ 
WD & Ooi et al. \cite{ooi:16} & 10  & 75 &  9.87 \\ 
WD & Zois et al. \cite{zois:16} & 5 & 75 & 6.02 \\ 
WD & Hafemann et al. \cite{hafemann:17} & 10 & 75  &  2.87 (0.42) \\ 
WD & Zois et al. \cite{zois:17} & 5 & 75 &  3.97\\ 
WD & Hafemann et al. (fine-tuned) \cite{hafemann:18} & 10 & 75 &  3.40 (1.08)\\ 
WD & Okawa \cite{okawa:18} & 10 & 75 &  6.40\\ 
WD & Zois et al. \cite{zois:18} & 5 & 75 &  3.52\\ 
WD & Zois et al. \cite{zois:19} & 10 & 75 &  1.37\\ 
WI & Zois et al. \cite{zois:19asymmetric} & 5 & 1 &  3.50\\ 
WI & Souza et al. \cite{souza:20} & 10 & 1 & 2.89 (0.13) \\ 
\hdashline
WI & $BPSO_{GV}$ & 10 & 1 & 2.48 (0.23) \\ 
\hline
\end{tabular}
\end{table}

From Tables \ref{tab:state_cedar} and \ref{tab:state_mcyt}, even operating in a transfer learning scenario the $BPSO_{GV}$ model was able to obtain low verification errors, comparable to the other state of the art models. 
Comparing with WD models, our WI $BPSO_{GV}$ outperforms half of the listed methods in CEDAR and is overpassed by only one in the MCYT dataset. Still, our approach has the advantage of being adaptable (since it is being used in a transfer learning context) and using only one classifier to perform the verification. 
For the WI comparison, in the CEDAR dataset, our approach outperformed better results than five of the eight models. When considering the MCYT dataset, our approach presents the best results.

\section{Conclusions}
\label{sec:conclusions}

In this work, we evaluated the use of BPSO-based feature selection for offline writer-independent handwritten signature verification. The optimization was conducted based on the minimization of the Equal Error Rate ($EER$) of the WI classifier in a wrapper mode. 

Results indicate the presence of redundant features in the dissimilarity space generated by the dichotomy transformation since the global validation ($BPSO_{GV}$) approach managed to obtain a better $EER$ using only 55\% of the total number of features.

Results also showed that not using an overfitting control scheme is worse than the partial validation ($BPSO_{PV}$) strategy, which in turn is worse than using the proposed $BPSO_{GV}$ approach. Thus, by using the global validation strategy it is possible to control the overfitting of the model and, thereby, improve the performance of the BPSO-based feature selection approach.

Finally, the space generated after feature selection can actually be used in a transfer learning context.
In a transfer learning scenario, the overfitting control has a bigger impact and the overfitting control scheme helps in having a feature representation that can generalize better across different datasets as it does not specialize too much on the optimization one.

Future works may include extending the analysis performed in this study to a multi-objective PSO, in order to minimize both the $EER$ and the number of features during the optimization.

\section*{Acknowledgment}

This work was supported by the FACEPE (Fundação de Amparo à Ciência e Tecnologia de Pernambuco),  CNPq (Conselho Nacional de Desenvolvimento Científico e Tecnológico) and the École de Technologie Supérieure (ÉTS Montréal).

\bibliographystyle{IEEEtran}
\bibliography{IEEEabrv,mybibliography}

\end{document}